\documentclass[letterpaper, 10 pt, conference]{ieeeconf}  %

\IEEEoverridecommandlockouts                              %

\usepackage{graphics} %
\usepackage{epsfig} %
\usepackage{times} %
\usepackage{amsmath} %
\usepackage{amssymb}  %
\usepackage{pgfplots}  %
\usepackage{float}  %
\usepackage{comment}
\usepackage[caption=false,font=footnotesize]{subfig}
\usepackage{fancyhdr}
\DeclareMathOperator*{\argmax}{arg\,max}
\DeclareMathOperator*{\argmin}{arg\,min}

\newcommand{\mc}[1]{\mathcal{#1}}
\newcommand{\mb}[1]{\mathbf{#1}}

\title{\LARGE \bf
Coverage Control in Multi-Robot Systems via Graph Neural Networks*
}
\author{Walker Gosrich$^{1}$, Siddharth Mayya$^{1}$, Rebecca Li$^{1}$, James Paulos$^{2}$, \\Mark Yim$^{1}$, Alejandro Ribeiro$^{3}$, Vijay Kumar$^{1}$%
\thanks{*This research was sponsored by the Army Research Lab through ARL DCIST CRA W911NF-17-2-0181.}%
\thanks{This research was sponsored by the National Science Foundation through award number 2036881.}%

\thanks{$^{1}$W. Gosrich, S. Mayya, R. Li, M. Yim, and V. Kumar are with the GRASP Laboratory, University of Pennsylvania, Philadelphia, PA, USA
        {\tt\small \{gosrich, mayya, robot, yim, kumar\}@seas.upenn.edu }}%

\thanks{$^{2}$ J. Paulos is with Treeswift, 3580 Indian Queen Ln., Philadelphia, PA {\tt\small jpaulos@seas.upenn.edu}}%
\thanks{$^{3}$A. Ribeiro is with the Department of Electrical and
Systems Engineering, University of Pennsylvania, Philadelphia, PA
{\tt\small aribeiro@seas.upenn.edu}}%

}

\begin{document}
\maketitle
\setlength{\headsep}{0.2in}
\thispagestyle{fancy}
\cfoot{}
\pagestyle{empty}

\begin{abstract}
This paper develops a decentralized approach to mobile sensor coverage by a multi-robot system. We consider a scenario where a team of robots with limited sensing range must position itself to effectively detect events of interest in a region characterized by areas of varying importance. Towards this end, we develop a decentralized control policy for the robots---realized via a Graph Neural Network---which uses inter-robot communication to leverage non-local information for control decisions. By explicitly sharing information between multi-hop neighbors, the decentralized controller achieves a higher quality of coverage when compared to classical approaches that do not communicate and leverage only local information available to each robot. Simulated experiments demonstrate the efficacy of multi-hop communication for multi-robot coverage and evaluate the scalability and transferability of the learning-based controllers.

\end{abstract}

\section{INTRODUCTION}
Coverage control, or distributed mobile sensing, considers the problem of distributing a team of robots in a region such that the likelihood of detecting events of interest is maximized~\cite{cortes2004coverage}. This is an extensively studied topic within the multi-robot systems literature (e.g., see survey paper~\cite{wang2011coverage} and references within), owing to its versatility in capturing a number of relevant multi-robot scenarios, such as surveillance~\cite{doitsidis2012optimal}, target tracking~\cite{khaledyan2019optimal, pimenta2009simultaneous}, and data collection via mobile sensor networks~\cite{zhong2011distributed}.  \par

While a diverse set of approaches have been developed to address the multi-robot coverage problem, e.g.,~\cite{liu2012dynamic, gao2020effective,elamvazhuthi2018pde}, an often used technique in the context of optimal sensor placement is the decentralized \textit{Lloyd's algorithm} ~\cite{du1999centroidal}. Lloyd's algorithm iteratively partitions the environment among the robots (typically using a \emph{Voronoi partition}~\cite{du1999centroidal}), and lets each robot monitor the area within its region of dominance~\cite{cortes2004coverage}. Given a partition, the coverage performance is then encoded by a locational reward function which evaluates the proximity of each robot to the points within its Voronoi cell. Lloyd's algorithm is the controller derived by computing the gradient of this reward function with respect to the robots' positions. \par 

\begin{figure}
	\centering 
	\includegraphics[width=0.95\linewidth]{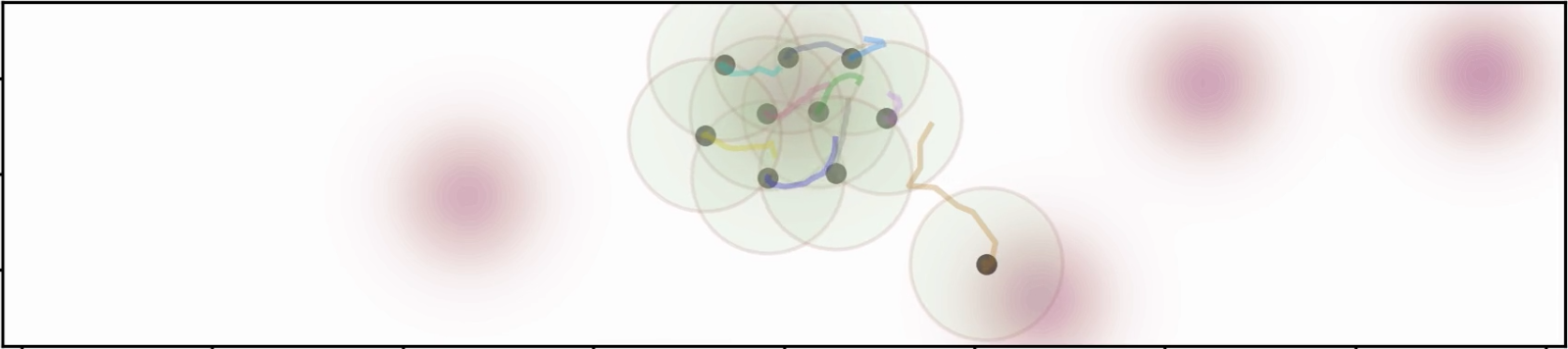}
	\caption{\label{fig:lloyds_fail_vignette} A team of 10 robots (depicted as green dots) with limited sensing ranges (visualized with disks) executing Lloyd's algorithm~\cite{cortes2004coverage} to cover an environment, where the shaded regions indicate areas of higher importance. The motion trails for each robot are depicted over 64 time steps. Lloyd's algorithm utilizes a purely gradient-based communication-free approach, and thus fails to drive the robots to an effective coverage configuration in this challenging scenario.} 
\end{figure}

In the context of multi-robot mobile sensor networks, the sensor placement problem is complicated by constraints on sensing and communication of the robots, typically approximated by limited-radius discs. These constraints limit the amount of information that is aggregated by each robot, and impact the approximated gradient that underpins Lloyd's algorithm ~\cite{cortes2005spatially}. To illustrate this point, Fig.~\ref{fig:lloyds_fail_vignette} considers a realistic scenario where robots with limited exteroceptive capabilities are tasked with covering a region where areas of unequal coverage importance are spread out over the region. These ``peaks" could represent targets which need to be monitored or areas with a high-likelihood of event occurrence. This scenario presents two challenges to Lloyd's algorithm: \emph{(i)} the region to be monitored by the robots is larger than what their sensing disks can cover, and \emph{(ii)} the robots are not initialized uniformly over the region (such as when initialized at a charging depot)~\cite{derenick2011energy}. As seen in Fig.~\ref{fig:lloyds_fail_vignette}, running Lloyd's algorithm in such a scenario results in a large part of the environment remaining uncovered and 4 out of 5 peaks being unmonitored. \par %

The scenario in Fig.~\ref{fig:lloyds_fail_vignette} demonstrates the potential value that \emph{inter-robot communication} could play in a coverage scenario---relevant information available to the exterior robots could be leveraged by other robots for improved decisions~\cite{yan2013survey}.  However, hand-designing such frameworks face the challenge of deciding \emph{what} information to communicate and \emph{how} to effectively use it~\cite{paulos2019decentralization}. Communication strategies developed for an application may end up being highly specific to the environment for which they're designed. \par

This paper develops a coverage control technique which leverages structured information-sharing between robots via a Graph Neural Network (GNN)~\cite{gama2020graphs} over the graph of robot nodes connected by undirected communication edges. In particular, the robots compute gradient information with respect to their local sensing region (informed by model-based approaches) but transform this information and communicate it with their neighbors to achieve non-myopic coordinated behaviors. This \emph{model-informed learning} solution enables relevant aspects of the coverage task to propagate through the network via communication among neighbors in the graph---while still retaining the advantage of a being a decentralized controller like Lloyd's algorithm. More specifically, we use \emph{imitation learning}~\cite{codevilla2018end} to train policies that map the decentralized state information available at each robot node to the actions generated by a clairvoyant expert controller, e.g.~\cite{tolstaya2020learning,gama2021graph}. \par %
GNNs provide an ideal framework for generalizing imitated behaviors to previously unseen coverage scenarios due to their ability to exploit symmetries in the graph topology of the robots (\emph{permutation equivariance}) and also tolerate changes in the network structure, which can occur as the robots move (\emph{stability
to perturbations})~\cite{gama2020stability}. Additionally, GNNs exhibit \emph{scalability} to graphs with larger numbers of robots than seen during training time~\cite{gama2020graphs}. \par

We present quantitative and qualitative results to showcase that the learned coverage policies outperform Lloyd's algorithm and cover a significantly higher number of peaks---which corresponds to a higher locational reward. These experiments are conducted for varying importance density distributions, starting robot configurations, sensing radii, and team sizes. Ablation studies explicitly demonstrate that the resulting policies leverage inter-robot communication for improved performance.

\section{RELATED WORK}~\label{sec:lit_rev}
Many variants of the coverage control problem have been considered in the literature, such as when certain regions in the environment are given a higher relative importance for coverage than others~\cite{lee2013controlled, santos2019decentralized}, when the robots have heterogeneous sensing capabilities~\cite{santos2018coverage}, and when different  sensor characteristics of the robots are considered~\cite{cortes2005spatially,laventall2009coverage}. In all cases, the resulting control laws are limited to gradient descent laws, which only leverage information available in the robots' immediate neighborhood.\par%
Many learning-based approaches to coverage have primarily focused on maximizing non-overlapping sensor footprints of the robots e.g.,~\cite{meng2021deep,pham2018cooperative} as well as scenarios where the regions of importance are incrementally mapped or learned through measurements~\cite{schwager2008consensus}. In~\cite{adepegba2016multi}, the authors develop a minimum-energy gradient descent law for robots with complex dynamics using an actor-critic reinforcement learning-based policy. However, the sensing range limitations of the robots are not considered.\par
In this paper, we leverage Graph Neural Networks (GNNs)~e.g., see~\cite{bruna2014spectral, ruiz2021graph,gama2018convolutional}, to design coverage control algorithms for robot teams by allowing individual robots to access non-local information via communication. GNNs have been shown to be effective at learning control policies in multi-robot systems, and fit well with the intuitive representation of multi-robot interactions with graphs~e.g.,~\cite{tolstaya2020learning, khan2021large}. Furthermore, as we discussed in the introduction, GNNs offer a wide range of advantages which are relevant in the context of multi-robot systems.

\section{PROBLEM SETUP: COVERAGE CONTROL}\label{sec:coverage_setup}
In this section, we formally introduce the sensor coverage scenario, present canonical solutions developed in the literature, and define the problem addressed in this paper. 

Consider a team of $N$ robots operating in a planar region, indexed by the set $\mc{N} = \{1,\ldots, N\}$, obeying single integrator dynamics:
\begin{equation}
    \mb{p}_{i,t+1} = \mb{p}_{i,t} + \mb{u}_{i,t},~ i\in\mc{N},
\end{equation}
where $\mb{p}_{i,t}\in\mathbb{R}^2$ and $\mb{u}_{i,t}\in\mc{U}\subseteq\mathbb{R}^2$ denote the position and velocity corresponding to robot~$i$ at discrete time $t = \{1,2,\ldots\}$.\par 

The robots are tasked with providing sensor coverage for a compact and convex planar region $\mc{D} \subset \mathbb{R}^2$, by positioning themselves at locations that maximize their ability to detect events of interest. The likelihood of occurrence of events of interest in the environment is encoded by a density field $\phi: \mc{D} \rightarrow \mathbb{R}$. We assume that this density field $\phi$ is fixed, but is unknown to the robots a priori. However, the robots are equipped with limited-range sensors which allow them to access values of the density field $\phi$ in the area covered by the sensor footprint. Let $r\in[0,\infty)$ denote the sensing radius for each robot in the team and $A_{i,t}(r) = \{\mb{q}\in\mc{D}~:~\|\mb{q} - \mb{p}_{i,t}\|\leq r\}$ denote the circular sensor footprint for robot $i$ at time $t$. Furthermore, we assume that the robots have access to the relative displacements to other robots in their neighborhood via local sensing and can communicate with any robot in the region (i.e., the communication radius is equal to or greater than the diameter of the region $\mc{D}$). \par 
For the purposes of coverage control, a common way to partition the environment is given as:
\begin{equation}\label{eqn:v_part}
    \mc{V}_{i,t} = \{\mb{q}\in\mc{D}~:~\|\mb{p}_{i,t} - \mb{q}\| \leq \|\mb{p}_{j,t} - \mb{q}\|, ~ \forall j\in\mc{N}\},
\end{equation}
denoted as a \emph{Voronoi} partition. Then, the quality of coverage provided by the team is encoded via the following locational function:
\begin{equation} \label{eqn:coverage-quality}
    \mc{J}(\mb{p}_{1,t},\ldots,\mb{p}_{N,t}) = -\sum_{i=1}^{N} \int_{\mc{V}_{i,t}} \|\mb{p}_{i,t} - \mb{q}\|^2\phi(\mb{q}) \mb{dq}
\end{equation}
which rewards the robots for being close to points in their Voronoi cell, as weighted by an importance density field $\phi$. Over a time horizon $t = 0,1,\ldots, T$, the objective of the coverage control algorithm is to choose control actions that drive the robots to maximize this coverage quality at the final time $T$:
\begin{align} \label{eqn:coverage_rew}
    \max_{\mb{u}_t, t\in\{0,\ldots,T\}}&~\mc{J}(\mb{p}_T) \\
    \text{s.t. } &\mb{p}_T = \mb{p}_0 + \sum_{t=0}^{T}\mb{u}_t \nonumber
\end{align}
where $\mb{p}_t = \left[\mb{p}_{1,t}^\top, \ldots, \mb{p}_{N,t}^\top\right]^\top$ and $\mb{u}_t = \left[\mb{u}_{1,t}^\top, \ldots, \mb{u}_{N,t}^\top\right]^\top$  denote the ensemble state and control inputs of the robot team, respectively. For the sake of notational simplicity, the rest of this paper drops the time index on variables unless explicitly specified. This implies that the computations occur as each discrete time $t$.

\subsection{Limited Range Lloyd's Algorithm}
As described in the introduction, Lloyd's algorithm consists of driving along the gradient of the coverage quality function described by~\eqref{eqn:coverage-quality} with respect to each robot's state~\cite{cortes2004coverage}. Since the sensor quality degradation encoded in $\mc{J}$ is continuous, this control law for robots with limited sensing radius $r$ can be computed based on a truncated computation of the centroid and mass of each Voronoi cell~\cite{cortes2005spatially}:
\begin{equation} \label{eqn:lloyds-action}
    \mathbf{u}_{i,r}^{\mc{L}} = 2m_{i}(r)(\mb{c}_{i}(r) - \mb{p}_{i}).
\end{equation}
where
\begin{align} \label{eqn:centroid}
    m_{i}(r) = &\int_{\mc{V}_{i}\cap A_{i}(r)}\phi(\mb{q})\mb{dq}, \\
    \mb{c}_{i}(r) = \frac{1}{m_{i}} &\int_{\mc{V}_{i}\cap A_{i}(r)}\mb{q}\phi(\mb{q})\mb{dq}.
\end{align}
This gradient descent law guarantees convergence to a local stationary point of the coverage quality function, which is also a \emph{centroidal Voronoi tessellation}, such that, for large enough episode length: $\mb{p}_{i,T} = \mb{c}_{i,T},\forall i\in\mc{N}$.
\subsection{Problem Definition} \label{sec:prob_def}
Consider $N$ robots with limited sensing range $r$, starting at locations $\mb{p}_0$ and tasked with covering a region $\mc{D}$ with an importance density field $\phi$. Let $\mc{J}_*^{\mc{L}}$ denote the coverage reward obtained by executing control law~\eqref{eqn:lloyds-action} until steady state is achieved. We are interested in synthesizing a decentralized feedback control policy $\mb{u}^{\mc{G}}$ which operates on the local information available to the robots and leverages inter-robot communication to obtain a coverage reward substantially higher than $\mc{J}_*^{\mc{L}}$.

\section{PROPOSED SOLUTION}
In this section, we present a learning-based approach to tackle the problem introduced in Section~\ref{sec:prob_def} by leveraging Graph Neural Networks (GNNs). First, we describe the graph model used for information sharing, followed by the GNN structure and the policy architecture. This section concludes with an imitation learning framework used to learn a decentralized policy that leverages communication to solve the coverage control problem.

\subsection{Inter-Robot Graph Design for Coverage}
We now formalize the graph model representing the multi-robot system and the corresponding graph signals that will be transformed into coverage actions in Section~\ref{sec:imit_learn}. Note that the computation of the Voronoi partitions in~\eqref{eqn:v_part} requires each robot to know the relative distance to ``adjacent" robots. This naturally induces a \emph{Delaunay graph} model in the robot team where nodes represent robots, and edges connect nodes which share a common boundary between their respective Voronoi cells~\cite{cortes2004coverage}. \par

In this paper, we superimpose an information exchange network on this same graph structure -- allowing robots to share information with relevant neighbors, as defined by the coverage task itself. Let $G = (V,\mathcal{E},W)$ represent a weighted and undirected Delaunay graph representing the robots and their neighbors as described above. We add scalar edge weights to the graph $w_{ij} \in W$ representing the normalized distance between agent $i$ and agent $j$:
\begin{equation} \label{eqn:edge-weights}
    w_{ij} = C||\mb{p}_i-\mb{p}_j||,~\forall (i,j) \in\mc{E}
\end{equation}
where $\|\cdot\|$ represents the $l_2$ norm, and $C$ is a suitably chosen normalization constant based on the size of the environment and the number of agents. 
The team-wide graph signal which is processed by the GNN is defined as 
\begin{equation}\label{eqn:def_graph_signal}
    \mb{x} = [\mb{v}_1, \mb{v}_2, \ldots, \mb{v}_N]^\top \in \mathbb{R}^{N \times 3}
\end{equation}
where the feature on each node ${\mb{v}_i} \in\mathbb{R}^3$ is the processed robot sensor information used by Lloyd's algorithm, given as:
\begin{equation} \label{eqn:node-features}
    \mb{v}_i = [(\mb{p}_i-\mb{c}_i(r))^\top, \; m_i(r)]^{\top}, \forall i\in\mc{N}
\end{equation}
where $\mb{p}_i, \mb{c}_i, m_i$ are defined in Section~\ref{sec:coverage_setup}.
\subsection{Graph Neural Networks}
Graph Neural Networks (GNNs) are a family of processing architectures that combine graph processing tools such as graph filters with machine learning principles (e.g. pointwise nonlinearities, layered architectures, and learning weights) to process information on graphs. In this paper, GNNs refer to a graph convolution operation followed by a nodewise non-linear operation---also known in the literature as \emph{graph convolutional neural networks}. \par
In each layer $l \in \{0,\ldots,L\}$, where $L$ denotes the total number of layers, the GNN processes a graph signal $\mb{z}_{l-1}$ with two successive operations: a \textit{ graph convolution} and a \textit{pointwise nonlinearity}. Each layer outputs a graph signal $\mb{z}_{l}$ to be processed by the next layer. The following equation represents layer $l$ of a GNN as a function $\Phi_l$:

\begin{equation}\label{eqn:gnn_l_th}
\mb{z}_{l} = \Phi_l(\mb{z}_{l-1}; \mb{S}, \mc{H}) = \sigma\left(\sum_{k=0}^K h_{l,k} \mb{S}^k\mb{z}_{l-1}\right),
\end{equation}
$\forall l\in\{1,\ldots,L\}$, where $\sigma(\cdot)$ applied to the graph convolution output is the \textit{pointwise nonlinearity}. \par 
The polynomial component of the GNN, $\sum_{k=0}^K h_{l,k} \mb{S}^k\mb{z}_{l-1}$, is the \textit{graph convolution}. It is defined as a polynomial on the \textit{adjacency matrix} $\mb{S} \in \mathbb{R}^{N\times N}$ multiplied by the learned weight $h_{l,k} \in \mathbb{R}$ corresponding to the $l^{th}$ layer and $k^{th}$ polynomial order. These weights are collected in the set $\mc H$. Leveraging the graph edge weights defined in~\eqref{eqn:edge-weights}, we consider a weighted version of the graph adjacency matrix, defined as: 
\begin{equation}
  \label{eqn:adjacency}
    \mb{S}_{ij} =
    \begin{cases}
      w_{ij}, & \text{if}\ (i,j) \in \mathcal{E} \\
      0, & \text{otherwise}.
    \end{cases}
\end{equation}
When multiplied with $\mb{z}_{l-1}$, it \textit{shifts} the graph signal, diffusing its feature vectors along the graph edges. \par 
Thus, the order of the polynomial in the graph convolution determines how far information travels via local communication: an order-$K$ graph convolution incorporates information into each node from its $K$\textit{-hop neighbors}---those nodes that lie $K$ edges away. In this paper, we consider at most second-order graph polynomials, and thus, assume that the accrued communication delay is negligible compared to the time scale of robot navigation. \par 
Together, these $L$ layers form the GNN, represented by the operation $\Phi(\mb{z}_{0};\mb{S},\mc{H})$ and composed of the layers described in~\eqref{eqn:gnn_l_th}. It is parameterized by the set of its weights $\mc{H}$, the graph over which it is applied $\mb{S}$, and the graph signal it is applied to $\mb{z}_{0}$. The output of the GNN is the output of the last layer $\mb{z}_{L}$.

\subsection{Policy Architecture}\label{sec:policy_arch}
In this section, we describe our specific modification of the classical GNN formulation described above, and apply it to the decentralized coverage problem. At each time step, we apply the GNN to the graph signal defined in~\eqref{eqn:def_graph_signal}: $\mb{z}_{0} = \mb{x}$. The GNN returns a vector of agent actions $\mb{u}^{\mc{G}}(\mb{x}; G,\mc{H})$. 

\begin{figure}
	\centering 
	\includegraphics[trim={1.0cm 2.0cm 4.0cm 1.5cm},clip,width=\linewidth]{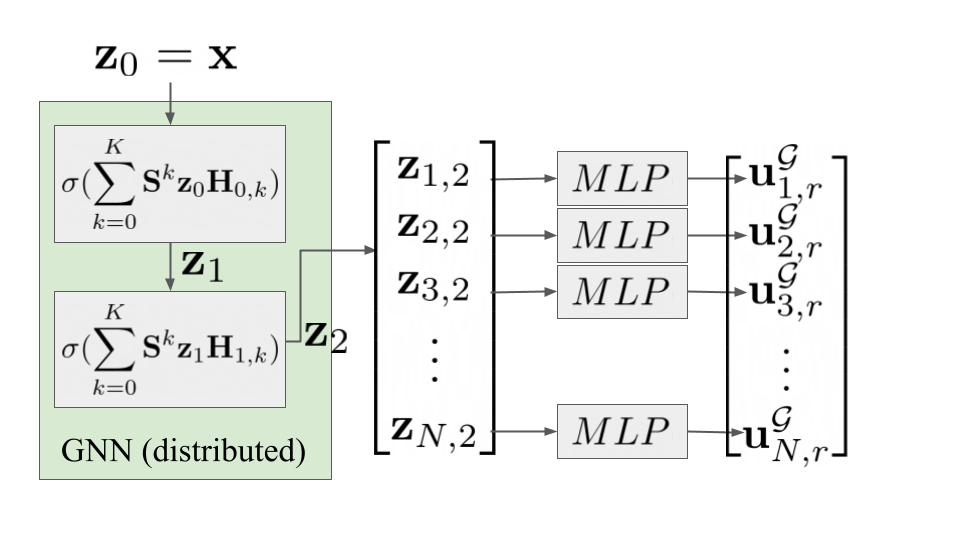}
	\caption{\label{fig:architecture} Policy architecture used in this paper for coverage control. The GNN is composed of two layers, and takes as input the node features defined in~\eqref{eqn:node-features} and outputs a feature vector $\mb{z}_{i,2}$ for each robot $i\in\mc{N}$. The MLP then acts upon each feature vector individually to compute the output action.}
\end{figure}

In order to enhance the expressivity of the graph convolution operations, we replace the scalar weights of the standard GNN $h_{l,k}$ with matrices $\mb{H}_{l,k}$. This permits separate scaling of each feature in the feature vector $\mb{v}_i$, and permits mixing among the features. By choosing $\mb{H}_{0,k} \in \mathbb{R}^{3\times \lambda}$ and $\mb{H}_{l,k} \in \mathbb{R}^{\lambda\times \lambda}, \forall l \in \{1,L\}$,  we can project the input feature vector $\mb{v}_i$ into a latent space of size $\lambda$. For each layer $l\in\{1,\ldots,L\}$, this is represented by the following modified operations compared to~\eqref{eqn:gnn_l_th}:
\begin{equation}\label{eqn:custom_gnn_l_th}
    \mb{z}_{l} = \Phi_l(\mb{z}_{l-1}; \mb{S}, \mc{H}) = \sigma\left(\sum_{k=0}^K \mb{S}^k\mb{z}_{l-1} \mb{H}_{l,k}\right).
\end{equation}
To transform the processed latent vectors into the action space of the robots, we append a decoder layer after the output of the GNN. The decoder consists of a two-layer multi-layer perceptron (MLP) with an input layer of size $\lambda$, hidden layer of size $32$ and output layer of size $p$, thus transforming each robot $i$'s GNN output---denoted as $\mb{z}_{i,L}$---into the control input, denoted by the vector $\mb{u}_{i,r}^{\mc{G}}(\mb{x}; G,\mc{H})$ where the subscript $r$ signifies that all computations were based on information collected from sensors of radius $r$. Note that in addition to the operations in~\eqref{eqn:custom_gnn_l_th} being decentralized, the MLP is evaluated separately at every node of the GNN (representing a robot), so it also represents a decentralized operation (see Fig.~\ref{fig:architecture}). The parameters chosen for the coverage controller were $ L=2$ and $\lambda = 64$, with the number of hops $K$ varied between $0$ and $2$ as will be demonstrated in Section~\ref{sec:experiment2}.\par

\subsection{Imitation Learning Framework}\label{sec:imit_learn}
As indicated in~\eqref{eqn:coverage_rew}, we are interested in maximizing the quality of coverage of the team \textit{at the final step}. Thus, at each time $t$, we are interesting in computing the control action for the robot team $\mb{u}^{\mc{G}}(\mb{x}_t; G_t,\mc{H})$ parameterized by the GNN weights $\mc{H}$. This can be posed as a search for the optimal weights $\mc{H}^*$ that maximize the reward at the final step $T$:
\begin{equation}
\mc{H}^* = \argmax_{\mc{H}}(\mathcal{J}(\mb{p}_T))
\end{equation}
\begin{equation*}
s.t. \; \;\mb{p}_T = \mb{p}_0 + \sum_{t=0}^T \mb{u}^{\mc{G}}(\mb{x}_t; G_t,\mc{H}).
\end{equation*}
To solve this problem, we employ an imitation learning strategy consisting of two primary components: an \textit{expert controller} that possesses centralized information, and a stochastic gradient descent strategy that modifies the GNN weights so that its output mimics that of the expert's.
\begin{figure}

    \centering
    \includegraphics[trim={0.0cm 1.6cm 1.0cm 0.2cm},clip,width=0.45\textwidth]{
	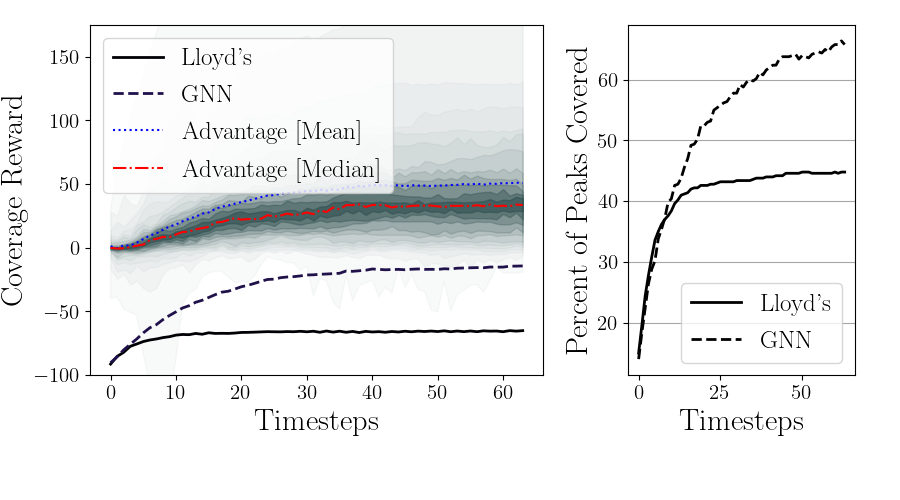}
     \caption{	\label{fig:reward_vs_time_difference}
 Experiment 1: the coverage reward \eqref{eqn:coverage-quality} vs time for 100 trials of 64 time steps. We show the mean reward for Lloyd's algorithm and for the GNN controller, as well as the mean and median \textit{reward advantage}---the amount by which the GNN outperformed Lloyd's during each trial. The shading indicates the statistical quantiles of the reward advantage over the trials. For the same experiments, the right plot shows the percentage of importance density field peaks that the controllers covered on average.}
\end{figure}

    ~
\begin{figure}
        \centering
        \includegraphics[trim={0.17cm 0.0cm 1.6cm 1.5cm},clip, width=0.45\textwidth]{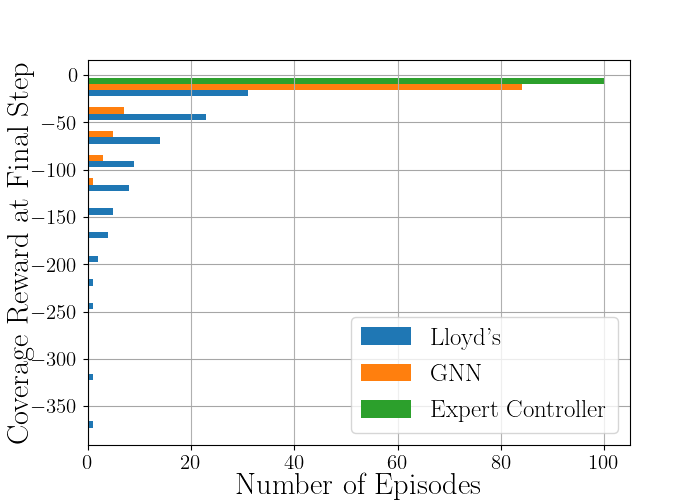}
        \caption{        \label{fig:triple_histogram_comparison}
Experiment 1: A histogram of the coverage reward performance in the final step (t=64) by three models: Lloyd's algorithm, the GNN controller, and the expert controller over 100 trials. The long-tailed distribution of Lloyd's performance highlights many scenario where gradient descent yields a low reward. Our GNN controller avoids these worst-case scenarios, and finds better final positions with more consistency.}
\end{figure}

\begin{figure}
        \centering
        \includegraphics[trim={0.3cm 0.3cm 1.8cm 1.5cm},clip,width=0.45\textwidth]{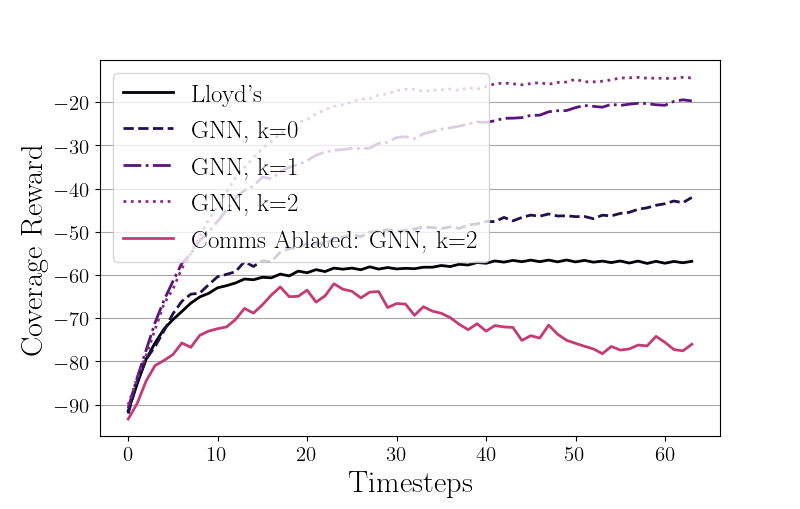}
        \caption{        \label{fig:reward_vs_time_no_comms}
The impact of communication on coverage performance over 100 trials for GNN models with zero (no communication, same as Lloyd's), one, and two hops of communication. We compare this with Lloyd's, and with the two-hop GNN controller tested with no communication. Models with greater communication range perform better, but all models outperform Lloyd's except the one with ablated communication.}
\end{figure}

\subsubsection{Expert Controller}
We designed a clairvoyant controller that leverages ``oracular" global information to search for robot configurations that yield a higher coverage reward when compared to Lloyd's algorithm. Towards this end, we designed an iterative sampling-based solution which repeatedly uses Lloyd's algorithm to find better robot configurations. Given a density function $\phi$, we conduct $M$ trials where the robots are randomly initialized in the domain and execute Lloyd's algorithm with an infinite sensor radius (generating control inputs $\mb{u}_{i,\infty}, \forall i\in\mc{N}$). 
We save the final agent position that results in the best final reward over all $M$ trials. For any initial conditions of the robots, the clairvoyant control then simply consists in driving in a straight line---using a control-limited proportional controller---towards this best final configuration to achieve the highest coverage reward. Robots are assigned to specific points in the configuration using the Hungarian assignment algorithm~\cite{Kuhn55thehungarian}.\par
We use this expert policy $\mb{u}^{E}$ to generate a dataset of state-action pairs for each timestep $\{(\mb{x}_t,\mb{u}_t^{E}) \; | \; t\in\{0,\ldots,T\}\}$ over thousands of episodes. Although the expert controller has access to global information, we only record the graph signal $\mb{x}$ (see \eqref{eqn:node-features}) and supporting graph $G$ (see \eqref{eqn:edge-weights}) that the GNN has access to during testing. The dataset is unordered, as only state-action pairs per timestep are recorded. For this work, we generated datasets of 64,000 episodes, each  of which was 64 time steps long.
 \subsubsection{MSE-Loss Training}
A training loss function is constructed with the aim of training the GNN to mimic the outputs of the clairvoyant controller described above. More specifically, we use stochastic gradient descent on the mean squared error (MSE) between the action predicted by the GNN $\mb{u}^{\mc G}$ and the action chosen by the expert controller $\mb{u}^{E}$:
\begin{equation}
\mc{H}^* = \argmin_\mc{H} \mathbb{E}^{\mb{u}^{E}}\left[||\mb{u}^{\mc{G}}(\mb{x}; G,\mc{H})-\mb{u}^{E}||^2\right]
\end{equation} 
where the expectation $\mathbb{E}^{\mb{u}^{E}}$ is taken with respect to the distribution of observed state as obtained by executing the expert policy. This results in a set of parameters $\mc{H}^*$ used during inference as described in Section~\ref{sec:policy_arch}. We trained the GNN for 256 epochs with a learning rate of 0.0001.

\section{EXPERIMENTAL RESULTS}\label{sec:exp}

\subsection{Experimental Platform}
We evaluated our trained GNN model in a simulated coverage environment. The environment is a long and narrow rectangle: 8 units wide ($w=8$) by 40 units long ($l=40$). It has a normalized importance density field represented by a Gaussian mixture model with 5 randomly placed peaks ($m=5$) of similar size.  

The number of agents and the agent sensing radius are fixed for each experiment, but the initial positions $\mb{p}_0$ are generated randomly in a clustered manner. As discussed in the introduction, such an initial configuration of robots represents a practical scenario and requires the robots to spread out in the environment in a way that would benefit from information exchange. Note that in the following experiments, both Lloyd's algorithm and the proposed GNN controller are exposed to identical state information; they have the same sensor radii and are always initialized with identical sets of random initial conditions.  %

\subsection{Results: Comparison to Lloyd's}

\subsubsection{Experiment 1} We compared the trained GNN controller against Lloyd's algorithm and the clairvoyant expert over 100 trials in the environment described above ($l=40, \; w=8, \; m=5$) with 10 agents, each with a sensing radius of 2 units ($N=10, \; r=2$). With these parameters, the sensor disks of the robots can cover at most 40\% of the total environment area.

The GNN controller achieved an average coverage reward (see \eqref{eqn:coverage-quality}) of $-14.3$ in the final step ($t=64$), a $50.9$ point improvement over Lloyd's average score of $-65.2$.  In this scenario, the expert controller used for training is near optimal, with an average final reward greater than $-1$. Figure \ref{fig:reward_vs_time_difference} (left) shows the mean coverage reward earned by the GNN and by Lloyd's vs. time over the 100 trials. We also show the \textit{reward advantage}, defined as the difference between the GNN's reward and Lloyd's reward throughout each trial. The statistical quantiles of the reward advantage are shaded on the figure. A histogram of this distribution of rewards at the final time ($t=64$) is shown for all three models in Fig.~\ref{fig:triple_histogram_comparison}. 

In Fig. \ref{fig:reward_vs_time_difference} (right) we show the same experiment, plotting the percentage of the importance density field's peaks that are covered by each controller. This intuitive metric further demonstrates the advantage of the GNN: the GNN covers on average more than three of the five peaks ($65\%$) by the end of the episode, whereas Lloyd's covers slightly more than two peaks ($45\%$).

\subsubsection{Experiment 2} \label{sec:experiment2} In this experiment, we isolate and highlight the role that \emph{communication} plays in the GNN controller's coverage performance. In the same environment ($l=40,\; w=8, \;N=10, \; m=5, \; r=2$), we trained GNN models with no communication ($k=0$), one-hop ($k=1$) and two-hop ($k=2$) communication, as defined in~\eqref{eqn:custom_gnn_l_th}. We additionally tested \textit{communication ablation} of the two-hop model, where we keep all conditions the same but remove the edges of the graph $G$, effectively disabling communication. Figure \ref{fig:reward_vs_time_no_comms} shows the results of this experiment. Trained agents with no communication ($k=0$) outperform Lloyd's slightly on the coverage reward, but largely learn to mirror Lloyd's algorithm. Training GNNs with communication significantly improves performance. Ablating communication moves the input data out of the distribution seen during training, and reduces GNN performance severely.

\begin{figure}
	\centering 
	\includegraphics[trim={0.2cm 0.3cm 0.2cm 0.2cm},clip,width=0.75\linewidth]{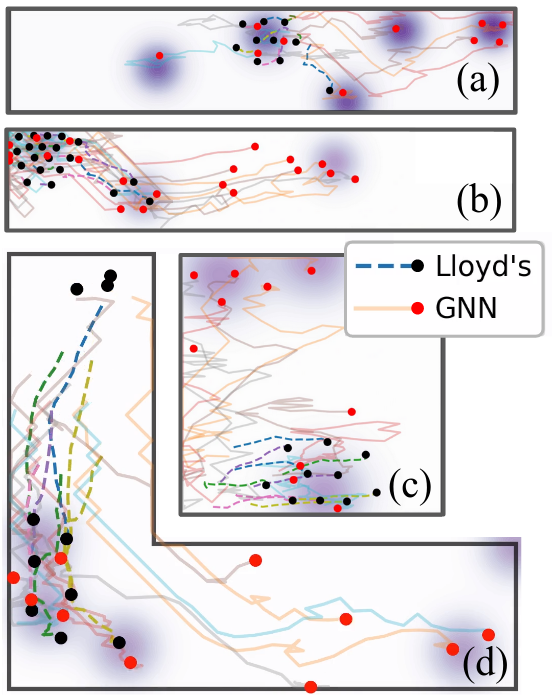}
	
	\caption{ Qualitative performance of the GNN controller compared to Lloyd's algorithm. The paths taken by the robots using both controllers are indicated by the dashed and solid lines. (a) shows the same episode as Fig.~\ref{fig:lloyds_fail_vignette}, and (b) demonstrates GNN performance with more robots ($N=20$) and fewer importance peaks ($m=3$). (c) and (d) show the GNN in environments different from the training environment in (a), including a non-convex environment. As seen, the GNN exhibits non-myopic coverage behaviors in all scenarios.}
	\label{fig:qualitative_comparison}
\end{figure}

The advantage of learned communication is demonstrated qualitatively in Fig.~\ref{fig:qualitative_comparison}, which shows snapshots selected from randomly generated trials. Whereas Lloyd's algorithm always descends the gradient of the coverage reward, the GNN has learned to use communication to travel through reward ``troughs,'' thus ending up in higher-reward states. \par 
It is noteworthy to see the ability of the GNN to generalize to new \emph{environments}. Even though it is trained only in the environment shown at the top of Fig.~\ref{fig:qualitative_comparison} ($l=40,\; w=8, \;N=10, \; m=5, \; r=2$) , it transfers well to different sized and shaped (even non-convex) environments, larger team sizes ($N=20$), and altered importance fields $(m=3$).

\subsection{Results: Transferability and Scalability}
We now demonstrate that our model-informed GNN controller performs well in a wide-range of test conditions, which are different from the conditions seen during training. More specifically, Experiment 3 evaluates the transferability of the GNN controller to robot teams with different sensor radii, and Experiment 4 shows generalization to larger teams of robots. The environment and density field parameters are identical to Experiments 1 and 2.
\subsubsection{Experiment 3} In Experiment 3, we trained 4 GNN models using datasets generated with different sensor radii $r=\{1,2,3,4\}$. This corresponds to the uniformly distributed sensor discs covering 10\% to 160\% (discs would overlap) of the environment. We test each model on $r=\{1,2,3,4\}$, and show the reward advantage over Lloyd's algorithm tested in the same conditions (see Fig.~\ref{fig:heatmap} (left)). Each row of the grid shows how a model trained on a given sensor radius transfers to other sensor radii. As expected, the reward advantage is always highest in the train conditions of the model. However, most of the generated models outperform Lloyd's algorithm even when the sensing radius is different from training. We also observe that models tested on sensing radii that are smaller than the training radius generally perform better than in the inverse case. %

\subsubsection{Experiment 4} 
\begin{figure}
	\centering 
	\includegraphics[trim={0.7cm 0.0cm 0.5cm 1.0cm},clip,width=\linewidth]{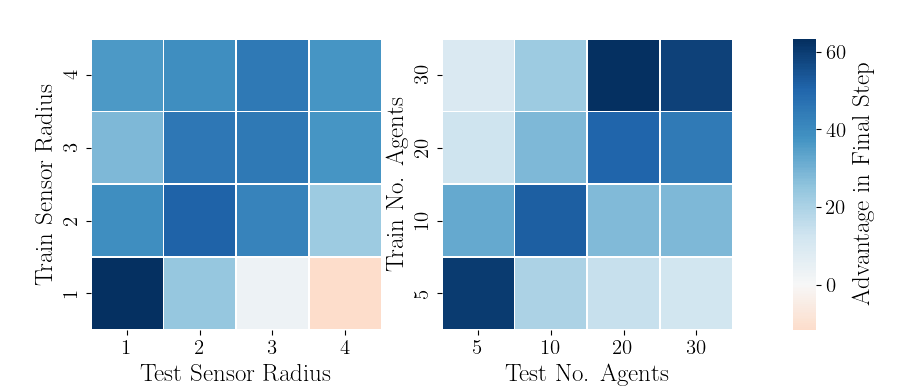}
	\caption{\label{fig:heatmap} Ensemble experiments demonstrating the flexibility of the GNNs: the colors indicate the reward advantage of the GNNs over Lloyd's. (left) Experiment 3 shows the transferability of GNNs to different sensor radii. The GNNs are strongest in the conditions in which they are trained, but are capable of outperforming Lloyd's in nearly all conditions evaluated. They tend to perform better when sensor radius is \textit{decreased}, rather than increased. (right) Experiment 4 demonstrates the scalability of GNNs to larger teams. They perform best in their train conditions, scale symmetrically to larger and smaller teams, and outperform Lloyd's in all cases.}
\end{figure}

In Experiment 4, we trained 4 GNN models on datasets with different numbers of robots: $N= \{5, 10, 20,  30\}$; and tested the models on the same range (see Fig.~\ref{fig:heatmap} (right)). Similar to Experiment 3, the models tend to perform the best when tested on their train conditions, and performance degrades as the number of agents is changed more significantly. However, all models in all test conditions outperformed Lloyd's algorithm, as shown by the reward advantage.

\section{CONCLUSION}

We have demonstrated how Graph Neural Networks can be leveraged to design decentralized coverage control algorithms that learn to leverage communication for improved performance. The GNN controller outperforms the canonical Lloyd's algorithm under a variety of testing conditions, including scenarios not seen during training. We demonstrate this capability with varying sensor radii, team sizes, and environment shapes as well as sizes. We use imitation learning to automatically synthesize such communicative controllers, and we show that the synthesized controllers' performance heavily relies upon this communication. This demonstrates the advantages of using GNNs in multi-robot control design, where communication is necessary, but designing bespoke communication strategies would be infeasible.

\section{ACKNOWLEDGEMENTS}
We thank Luana Rubini Ruiz for extremely helpful discussions regarding the training and architecture design for GNNs.

\addtolength{\textheight}{-6cm}   %

\bibliographystyle{unsrt}
\bibliography{references}

\end{document}